\documentclass[letterpaper]{article} %

\usepackage[]{aaai23}  %
\usepackage{times}  %
\usepackage{helvet}  %
\usepackage{courier}  %
\usepackage[hyphens]{url}  %
\usepackage{graphicx} %
\usepackage{natbib}  %
\usepackage{caption} %
\usepackage{algorithm}
\usepackage{algorithmic}

\usepackage{newfloat}
\usepackage{listings}
\DeclareCaptionStyle{ruled}{labelfont=normalfont,labelsep=colon,strut=off} %
\floatstyle{ruled}
\newfloat{listing}{tb}{lst}{}
\floatname{listing}{Listing}
\usepackage{amsmath}
\usepackage{amssymb}
\usepackage{mathtools}
\usepackage{bbm}
\usepackage{bm}
\usepackage{scalerel} %

\usepackage{booktabs}
\usepackage{multirow}
\usepackage{subcaption}
\usepackage{adjustbox}
\usepackage{microtype}
\usepackage{algorithm}  
\usepackage{algorithmic}  

\usepackage[justification=justified, skip=1pt]{caption}

\usepackage[inline]{enumitem}
\usepackage[capitalize]{cleveref}

\usepackage{tikz}
\usetikzlibrary{arrows.meta}
\usetikzlibrary{backgrounds}
\usetikzlibrary{positioning, fit, mindmap, trees, calc, tikzmark, shapes}
\usetikzlibrary{shapes.arrows, fadings, automata, tikzmark, decorations.pathreplacing, patterns}
\usetikzlibrary{arrows.meta}
\usetikzlibrary{intersections}
\usepackage{tkz-euclide}

\usepackage{pgfplots}
\usepgfplotslibrary{groupplots}
\pgfplotsset{compat=1.14}

\usepackage{sistyle}
\SIthousandsep{,}

\usepackage{soul}
\setstcolor{red}

\usepackage{xcolor}
\usepackage{diagbox}

\definecolor{tea_green}{RGB}{214, 234, 193}
\definecolor{hint_green}{RGB}{226,246,209}
\definecolor{Madang}{RGB}{190,235,159}
\definecolor{yellow_green}{RGB}{198,222,119}
\definecolor{link_water}{RGB}{221, 232, 250}
\definecolor{celestial_blue}{RGB}{52, 152, 219}
\definecolor{shakespeare}{RGB}{85, 154, 193}
\definecolor{buttermilk}{RGB}{255,242,174}
\definecolor{chardonnay}{RGB}{250,196,114}
\definecolor{rajah}{RGB}{253,180,98}
\definecolor{fog}{RGB}{213, 193, 234}
\definecolor{melon}{RGB}{254,191,181}
\definecolor{sundown}{RGB}{249, 180, 181}
\definecolor{mona_lisa}{RGB}{246,152,134}
\definecolor{salmon}{RGB}{242,131,107}

\definecolor{saltpan}{RGB}{238, 243, 232}
\definecolor{aqua_spring}{RGB}{232, 243, 232}
\definecolor{tea_green}{RGB}{214, 234, 193}
\definecolor{Madang}{RGB}{190,235,159}
\definecolor{fringy_flower}{RGB}{194, 234, 193}
\definecolor{aero_blue}{RGB}{193, 234, 213}
\definecolor{pixie_green}{RGB}{183,214,170}
\definecolor{french_pass}{RGB}{195,232,246}
\definecolor{ice_cold}{RGB}{169,232,220}
\definecolor{pale_turquoise}{RGB}{172,240,242}
\definecolor{cruise}{RGB}{179,226,205}
\definecolor{sail}{RGB}{163,205,235}
\definecolor{spindle}{RGB}{179,205,227}
\definecolor{link_water}{RGB}{221, 232, 250}
\definecolor{periwinkle}{RGB}{203,213,232}
\definecolor{zanah}{RGB}{220, 233, 213}
\definecolor{frostee}{RGB}{217, 231, 214}
\definecolor{opal}{RGB}{199, 221, 211}
\definecolor{jet_stream}{RGB}{188, 214, 210}
\definecolor{skeptic}{RGB}{153, 187, 167}
\definecolor{hint_green}{RGB}{226,246,209}
\definecolor{snow_flurry}{RGB}{230,245,201}
\definecolor{surf_crest}{RGB}{205,230,208}
\definecolor{yellow_green}{RGB}{198,222,119}
\definecolor{cream}{RGB}{255,255,204}
\definecolor{pale_prim}{RGB}{255,255,179}
\definecolor{spring_sun}{RGB}{242,243,195}
\definecolor{portafino}{RGB}{245,237,160}
\definecolor{buttermilk}{RGB}{255,242,174}
\definecolor{cream_brulee}{RGB}{255, 229, 151}
\definecolor{dairy_cream}{RGB}{254,226,189}
\definecolor{champagne}{RGB}{254,217,166}
\definecolor{chardonnay}{RGB}{250,196,114}
\definecolor{manhattan}{RGB}{226,180,125}
\definecolor{rajah}{RGB}{253,180,98}
\definecolor{early_dawn}{RGB}{252,243,218}
\definecolor{egg_shell}{RGB}{238, 234, 215}
\definecolor{selago}{RGB}{243, 232, 243}
\definecolor{quartz}{RGB}{219,223,238}
\definecolor{fog}{RGB}{213, 193, 234}
\definecolor{languid_lavender}{RGB}{222,203,228}
\definecolor{watusi}{RGB}{254,221,207}
\definecolor{coral_andy}{RGB}{243,204,205}
\definecolor{cosmos}{RGB}{248,209,210}
\definecolor{melon}{RGB}{254,191,181}
\definecolor{azalea}{RGB}{234, 193, 194}
\definecolor{beauty_bush}{RGB}{235, 185, 179}
\definecolor{sundown}{RGB}{249, 180, 181}
\definecolor{mona_lisa}{RGB}{246,152,134}
\definecolor{salmon}{RGB}{242,131,107}

\definecolor{summer_sky}{RGB}{58, 151, 233}
\definecolor{chateau_green}{RGB}{72, 179, 96}
\definecolor{matisse}{RGB}{25, 104, 167}
\definecolor{allports}{RGB}{31, 106, 125}
\definecolor{sun_shade}{RGB}{255, 144, 68}
\definecolor{flamingo}{RGB}{237, 88, 85}
\definecolor{studio}{RGB}{128, 91, 160}

\definecolor{maya_blue}{RGB}{102, 204, 255}
\definecolor{feijoa}{RGB}{178,223,138}
\definecolor{sushi}{RGB}{117, 168, 47}
\definecolor{norway}{RGB}{158, 194, 132}
\definecolor{japanese_laurel}{RGB}{53, 116, 40}
\definecolor{see_green}{RGB}{161,228,195}
\definecolor{monte_carlo}{RGB}{135,204,194}
\definecolor{granny_smith_apple}{RGB}{150,214,150}
\definecolor{moss_green}{RGB}{170,216,176}
\definecolor{chateau_green}{RGB}{72, 179, 96}
\definecolor{opal}{RGB}{164,207,190}
\definecolor{acapulco}{RGB}{117, 170, 148}
\definecolor{viridian}{RGB}{55, 137, 122}
\definecolor{amazon}{RGB}{56, 123, 84}
\definecolor{asparagus}{RGB}{123, 160, 91}
\definecolor{fruit_salad}{RGB}{91, 160, 94}
\definecolor{puerto_rico}{RGB}{72, 179, 150}
\definecolor{mountain_meadow}{RGB}{0, 163, 136}
\definecolor{matisse}{RGB}{25, 104, 167}
\definecolor{allports}{RGB}{31, 106, 125}
\definecolor{astral}{RGB}{55, 111, 137}
\definecolor{spring_leaves}{RGB}{46, 83, 117}
\definecolor{biscay}{RGB}{44, 62, 80}
\definecolor{midnight}{RGB}{0, 29, 50}
\definecolor{amethyst}{RGB}{153, 102, 204}
\definecolor{studio}{RGB}{128, 91, 160}
\definecolor{tapestry}{RGB}{194, 109, 132}
\definecolor{atomic_tangerine}{RGB}{255, 153, 102}
\definecolor{amber}{RGB}{255, 191, 0}
\definecolor{casablanca}{RGB}{244, 178, 84}
\definecolor{california}{RGB}{233, 140, 58}
\definecolor{tomato}{RGB}{255, 97, 56} 
\definecolor{alizarin}{RGB}{233, 58, 64}

\definecolor{linen}{RGB}{251, 239, 227}
\definecolor{double_pearl_lusta}{RGB}{253, 242, 208}
\definecolor{oasis}{RGB}{253, 242, 208}
\definecolor{milan}{RGB}{255, 254, 169}
\definecolor{texas}{RGB}{245, 232, 123}
\definecolor{maize}{RGB}{249, 212, 156}

\definecolor{turmeric}{RGB}{211, 178, 76}
\definecolor{saffron}{RGB}{249,193,62}
\definecolor{my_sin}{RGB}{255, 176, 59}
\definecolor{tree_poppy}{RGB}{246, 154, 27}
\definecolor{jaffa}{RGB}{240, 131, 58}
\definecolor{crusta}{RGB}{254, 127, 44}
\definecolor{tahiti_gold}{RGB}{223, 102, 36}
\definecolor{outrageous_orange}{RGB}{255, 100, 45}
\definecolor{safety_orange}{RGB}{254, 106, 0}

\definecolor{azalea}{RGB}{251, 196, 196}
\definecolor{oyster_pink}{RGB}{238,206,205} 
\definecolor{coral_candy}{RGB}{242,208,205} 
\definecolor{baby_pink}{RGB}{246, 194, 192}
\definecolor{petite_orchid}{RGB}{223, 157, 155}
\definecolor{apricot}{RGB}{241,140,122}
\definecolor{NY_pink}{RGB}{228,136,113}
\definecolor{carmine_pink}{RGB}{231, 76, 60}
\definecolor{deep_carmine_pink}{RGB}{236, 50, 67}

\definecolor{wewak}{RGB}{244, 143, 150}
\definecolor{light_coral}{RGB}{244, 127, 123}
\definecolor{bittersweet}{RGB}{255,111,105}
\definecolor{carnation}{RGB}{245, 80, 86}
\definecolor{flamingo}{RGB}{237, 88, 85}
\definecolor{sunset_orange}{RGB}{242,89,75}
\definecolor{ku_crimson}{RGB}{243, 0, 25}
\definecolor{amaranth}{RGB}{234,46,73}
\definecolor{valencia}{RGB}{214, 87, 70}
\definecolor{chilean_fire}{RGB}{215, 87, 44}
\definecolor{mexican_red}{RGB}{170, 41, 37}

\definecolor{napa}{RGB}{163, 154, 137}

\definecolor{athens_gray}{RGB}{236, 240, 241}
\definecolor{gallery}{RGB}{240,240,240}
\definecolor{mercury}{RGB}{230,230,230}
\definecolor{platinum}{RGB}{228,228,228}
\definecolor{silver}{RGB}{191,191,191}
\definecolor{aluminum}{RGB}{153,153,153}
\definecolor{ship_gray}{RGB}{77,77,77}
\definecolor{tuatara}{RGB}{67, 67, 67}

\definecolor{malibu}{RGB}{110, 180, 240}
\definecolor{celestial_blue}{RGB}{52, 152, 219}
\definecolor{curious_blue}{RGB}{41, 128, 185}
\definecolor{french_blue}{RGB}{0, 112, 182}
\definecolor{matisse}{RGB}{25, 104, 167}
\definecolor{shakespeare}{RGB}{85, 154, 193}
\definecolor{seagull}{RGB}{128,177,211}
\definecolor{jelly_bean}{RGB}{45, 126, 150}
\definecolor{venice_blue}{RGB}{87, 135, 105}
\definecolor{boston_blue}{RGB}{68, 147, 161}

\definecolor{turquoise}{RGB}{41,217,194}
\definecolor{java}{RGB}{2,190,196}
\definecolor{riptide}{RGB}{141,211,199}
\definecolor{mountain_meadow}{RGB}{0, 163, 136}
\definecolor{free_speech_aquamarine}{RGB}{0, 156, 114}

\definecolor{cosmic_latte}{RGB}{222, 247, 229}
\definecolor{chinook}{RGB}{163, 232, 178}
\definecolor{padua}{RGB}{121, 189, 143}
\definecolor{ocean_green}{RGB}{79, 176, 112}
\definecolor{pastel_green}{RGB}{107, 227, 135}
\definecolor{chateau_green}{RGB}{69, 191, 85}
\definecolor{RoyalBlue}{RGB}{69, 191, 85}
\definecolor{pigment_green}{RGB}{0, 175, 79}
\definecolor{fern}{RGB}{101,197,117}
\definecolor{killarney}{RGB}{56, 113, 66}

\newcommand{\argmin}{\operatornamewithlimits{arg\,min}}

\newcommand{\ie}{i.e., }
\newcommand{\eg}{e.g., }

\title{\textit{LegoNet}: A Fast and Exact Unlearning Architecture}

\author{Sihao Yu, Fei Sun, Jiafeng Guo$^*$, Ruqing Zhang, Xueqi Cheng\\
University of Chinese Academy of Sciences, Beijing, China\\
CAS Key Lab of Network Data Science and Technology, Institute of Computing Technology, \\ 
Chinese Academy of Sciences, Beijing, China\\
{\tt\small \{yusihao,sunfei,guojiafeng,zhangruqing,cxq\}@ict.ac.cn}
}

\begin{document}

\maketitle

\begin{abstract}

Machine unlearning aims to erase the impact of specific training samples upon deleted requests from a trained model.
Re-training the model on the retained data after deletion is an effective but not efficient way due to the huge number of model parameters and re-training samples. 
To speed up, a natural way is to reduce such parameters and samples.
However, such a strategy typically leads to a loss in model performance, which poses the challenge that increasing the unlearning efficiency while maintaining acceptable performance.
In this paper, we present a novel network, namely \textit{LegoNet}, which adopts the framework of ``fixed encoder + multiple adapters''.
We fix the encoder~(\ie the backbone for representation learning) of LegoNet to reduce the parameters that need to be re-trained during unlearning.
Since the encoder occupies a major part of the model parameters, the unlearning efficiency is significantly improved.
However, fixing the encoder empirically leads to a significant performance drop.
To compensate for the performance loss, we adopt the ensemble of multiple adapters, which are independent sub-models adopted to infer the prediction by the encoding~(\ie the output of the encoder).
Furthermore, we design an activation mechanism for the adapters to further trade off unlearning efficiency against model performance.
This mechanism guarantees that each sample can only impact very few adapters, so during unlearning, parameters and samples that need to be re-trained are both reduced.
The empirical experiments verify that LegoNet accomplishes fast and exact unlearning while maintaining acceptable performance, synthetically outperforming unlearning baselines.

\end{abstract}

\section{Introduction}

\begin{figure*}[tb]
\centering
\includegraphics[width=0.85\linewidth]{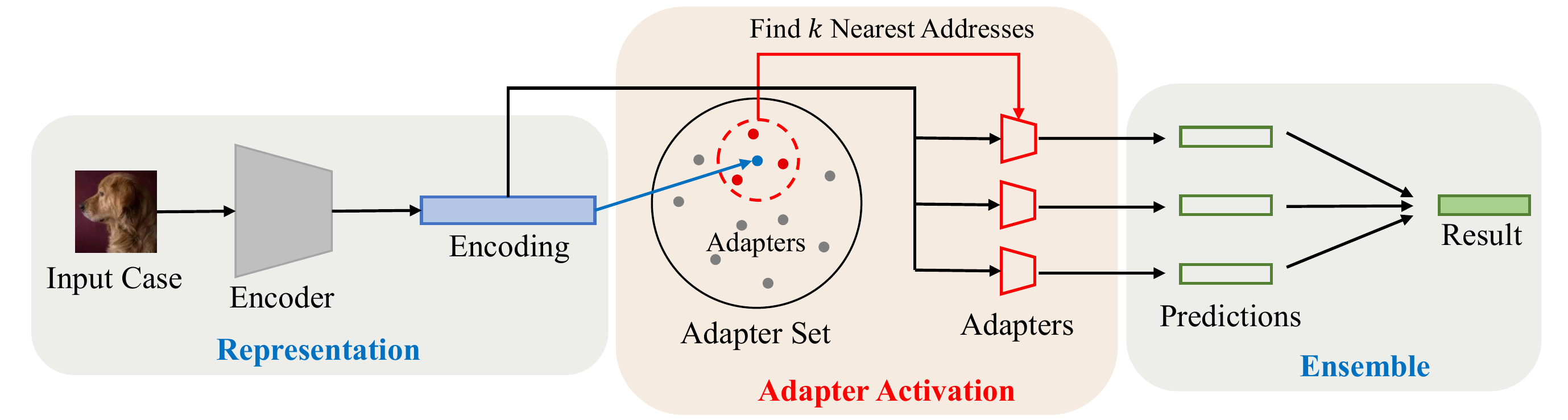}
\caption{Overview of LegoNet. The feedforward computation of LegoNet has three main processes, \ie representation, adapter selection, and ensemble. The input case is encoded into the encoding at first. Then, according to the encoding, $k$~($k=3$ in the picture) adapters are selected from the adapter set. Each adapter has independent parameters and makes one candidate prediction when receiving the input encoding. At last, the candidate predictions of activated adapters are aggregated to the final result.}
\label{fig:framework}
\end{figure*}

While machine learning brings great benefits, it also brings critical privacy concerns in recent years due to the unintended memorization of the training data which might contain users' sensitive information~\cite{Carlini:SEC19:Secret}, \eg users' behavior data used to train recommender systems~\cite{zhang2021membership} or personal information in the corpus for large natural language models~\cite{carlini2021extracting, zanella2020analyzing}.
To protect the individual's privacy, multiple data protection regulations, e.g., GDPR in the European Union, have enacted the ``\textit{right to be forgotten}'' which requires the service providers that process personal data to delete user data upon request~\cite{Alessandro2013The}.
However, only deleting data is not enough, considering that users' information could be leaked from the trained models~\cite{2017PrivacyRisk,Carlini:SEC19:Secret,Shokri:SP17:Membership,zhang2021membership,truex2019demystifying}.
To address this issue, machine unlearning has been proposed to erase the impact of specific training samples from a trained model upon deleted~(\ie unlearned) requests.

Numerous unlearning methods try to eliminate the impact of the unlearned samples by continually optimizing~(fine-tuning) the trained model's parameters~\cite{Guo:ICML20:Certified,golatkar2020eternal,Izzo:AISTATS21:Approximate}.
However, such methods typically can not guarantee that the impact from unlearned samples is completely removed, especially for the neural models.
To protect privacy, exact unlearning is required.
A naive way is to re-train the model from scratch using the retained training data.
However, such a fully re-training strategy has a critical issue in efficiency since all model parameters should be re-trained on all re-trained samples once a sample is required to be unlearned~\cite{bourtoule2021machine,chen2022recommendation,chen2021graph}.
Previous works have tried to reduce the samples that participate in re-training to accelerate the unlearning process~\cite{bourtoule2021machine,chen2022recommendation,chen2021graph}.
As a representative, SISA splits the data into multiple disjoint shards for training multiple independent sub-models one to one~(each sub-model adopts the same network as the original model unlearned by naive re-training).
When unlearning a sample, it only needs to re-train the corresponding sub-model on the retained samples in the shard to which the unlearned sample belongs.
As the number of shards increases, the samples involved in re-training are reduced, but the performance empirically drops significantly.
To maintain acceptable performance, the degree to which data can be sharded is severely limited.
Therefore, the SISA-based methods are still not efficient enough, especially when the amount of training data is large.

To further improve the unlearning efficiency, we propose a new idea that aims to reduce the number of parameters for re-training.
Inspired by continual learning~\cite{2021Kadapter,aljundi2017expert}, a feasible solution is to segment the parameters and reduce the impact region of the samples on the parameters.
Generally, a neural model has a large backbone to learn representations, and then adopts a simple network, such as a linear layer, to make predictions based on the representations.
While the representation learning of the backbone does not necessarily require the use of the task's training data.
It indicates that the backbone can get rid of re-training.
Then, the unlearning efficiency will be significantly improved.

Follow this idea, we present a novel neural network, named \textit{LegoNet}, as shown in \cref{fig:framework}.
Specifically, LegoNet consists of a fixed encoder~(\ie backbone) for representation and $n$ isolated adapters for determining the predictions.
Here, the encoder has been pre-trained on external data~(\eg a large generic dataset such as ImageNet~\cite{2015ImageNet}), which ensures the model has enough ability for representation.
Since the fixed encoder occupies most parameters of LegoNet, the re-trained parameters during unlearning are significantly reduced.
However, empirically, the fixed encoder also leads to a loss in model performance.
Here, LegoNet adopts the ensemble of multiple adapters to largely recover this performance loss.

Moreover, to further trade off model performance against unlearning efficiency, we design the adapter activation mechanism.
As the encoder maps the input data from the sample space to the encoding space, we assign preset keys for adapters to represent their address in the encoding space.
By this approach, we build the distance property between the adapters and the samples.
For each sample, $k$~($k\ll n$) nearest adapters will be activated and adopted to train or infer.
The ensemble of $k$ activated adapters ensures acceptable performance.
Moreover, the impact of any training sample is controlled in $k$ definite adapters.
Therefore, when unlearning a specific sample, LegoNet only needs to re-train $k$ impacted adapters to achieve exact unlearning.
Overall, such a mechanism further speeds up re-training in two ways.
On the one hand, although the number of adapters' parameters is already small, we further reduce the number of impacted adapters that need to be re-trained.
On the other hand, the number of samples that should be adopted to re-train the impacted adapters is reduced, since the adapter can only be impacted by the nearby training samples in the encoding space.
Since the number of both parameters and samples for re-training is effectively reduced, the unlearning speed of LegoNet is faster than that of SISA-based methods which only reduces the number of re-training samples.

To verify the exact performance of our unlearning approach, we conduct image classification experiments on CIFAR-10/100~\cite{2008cifar}.
Experiments present that our model efficiently accomplishes exact unlearning and keeps the performance on the test set.
Moreover, we compare LegoNet with SISA through ablation experiments to analyze the effects and advantages of our differences with SISA.
Finally, we conduct extensive experiments with different hyperparameters for LegoNet to further analyze how our approach trades off model performance against unlearning speed.

\section{Related Works}
\label{sec:rel}

Machine unlearning~\cite{bourtoule2021machine,Neel:ALT21:Descent,2021AdaptiveMU} refers to a process that aims to remove the impact of the deleted samples of training data from a trained model.
Such a process is also named as \textit{decremental learning}~\cite{Cauwenberghs:NIPS00:Incremental,Karasuyama:NIPS09:Multiple}, \textit{selective forgetting}~\cite{golatkar2020eternal} or \textit{data removal/deletion}~\cite{Guo:ICML20:Certified,Ginart:NIPS19:Making} in prior works.
From the certainty of removal, existing methods of machine unlearning can be divided into exact unlearning and approximate unlearning.

\textbf{Exact unlearning} ensures that the unlearned data is completely deleted from the trained model.
Previous studies typically achieve the exact data deletion at a cheaper computing cost than fully re-training for simple models or under specific conditions~\cite{Cauwenberghs:NIPS00:Incremental,cao2015towards,Schelter2020AmnesiaM}.
For example, leave-one-out cross-validation speed up the exact unlearning for SVMs (Support Vector Machines)~\cite{Cauwenberghs:NIPS00:Incremental,Karasuyama:NIPS09:Multiple}, provably efficient data deletion is designed for $k$-means clustering~\cite{Ginart:NIPS19:Making}, and
fast data deletion for Na\"ive Bayes is achieved with the assumption of statistical query learning (training data is in a decided order)~\cite{cao2015towards}.
More generally, SISA (Sharded, Isolated, Sliced, and Aggregated)~\cite{bourtoule2021machine}  proposes a representative framework that supports partial re-training, based on the idea of ``divide and conquer''.
Specifically, its main steps include: (1) divide the training data into some disjoint shards; (2) copy multiple models and train them on different shards independently; (3) aggregate the results of all models for the final inference.
Then, SISA exactly re-trains the impacted sub-model, whose shard contains the removal data, for unlearning.
Compared with fully re-training a model, re-training each model on tiny shards greatly improves the efficiency.
Following this idea, an improved sharding algorithm is presented for the unlearning of the graph~\cite{chen2021graph}. 
The core of SISA is data segmentation, which reduces the amount of data received by each sharded model for re-training.
However, the number of parameters for each re-training is still huge, which severely affects the efficiency.
Only reducing the amount of data has limited potential for acceleration, when SISA increases the number of data shards, the performance typically drops significantly.
In this paper, we try to reduce both parameters and samples for re-training to further speed up the unlearning.

\textbf{Approximate unlearning} relaxes the requirement for exact deletion~\cite{Guo:ICML20:Certified,golatkar2020eternal,Izzo:AISTATS21:Approximate}, \ie it only accomplishes that the removed data is statistically unlearned with the guarantee that the unlearned model cannot be distinguished from an exact deletion model~\cite{Guo:ICML20:Certified}.
Such approximate methods typically adopt gradient-based strategies to remove the influence of deleted samples from the trained model~\cite{Neel:ALT21:Descent}.
Following the idea of gradient update, different Newton's methods are proposed to approximate re-training for convex models, \eg linear regression, logistic regression, and the last fully connected layer of a neural network~\cite{Guo:ICML20:Certified,golatkar2020eternal,Golatkar:ECCV20:Forgetting}.
Since the core is to eliminate the influence of specified samples from the trained model, the gradient updates also can be decided according to the influence functions~\cite{Izzo:AISTATS21:Approximate}.
Compared with exact unlearning, approximate unlearning methods are usually more efficient.
However, such gradient-based methods are limited by the condition of ``convex'', thus they are hard to apply to non-convex models like deep neural networks. 
Moreover, such methods can not guarantee that the impacts of deleted data are completely removed.
Thus, their security is insufficient from a privacy protection point of view.

\section{LegoNet}
\label{sec:LegoNet}
In this section, we introduce our network, \ie \textit{LegoNet}, and the corresponding unlearning process.
First, we introduced the network architecture and the feedforward calculation of LegoNet.
Next, we introduce the training process and the unlearning process of LegoNet.
Finally, we compare LegoNet with SISA to further present and analyze the advantages of our method.

\subsection{Architecture and Feedforward Calculation}
\label{subsec:framework}
LegoNet is composed of a fixed encoder and $n$ isolated adapters.
The encoder, as the backbone of LegoNet, is a feature extractor adopted for representing the input samples.
Specifically, it maps the samples to the encoding space.
Here, the choice of the encoder's architecture is relatively free.
To deal with different kinds of inputs, we can adopt different architectures, such as BERT~\cite{vaswani2017attention} for text tasks or ResNet~\cite{2016Identity} for image tasks.
Each adapter contains a preset key to represent its address in the encoding space, and an independent sub-model adopted for prediction based on the output of the encoder.
Typically, the sub-model of each adapter only requires a simple architecture, such as a linear layer.
To facilitate the following description, we adopt symbols for formulation. \cref{symbol} gives the definitions of the main symbols used in this paper.

For inferring an input case $x$, the encoder firstly encodes $x$ to the encoding as $\bm{e}=\mathrm{En}(x)$.
This encoding determines the activation of the adapter.
Specifically, for $j=1$ to $n$, we can calculate the distance from the sample $x$ to the adapter $a_j$ as:
\begin{equation}
\label{equ:du-activate}
    \delta(x, a_j)=\Vert \mathrm{En}(x)-\bm{K}_j \Vert =\Vert \boldsymbol{e}-\bm{K}_j\Vert,
\end{equation}
where $\Vert\cdot \Vert$ denotes the norm~(\eg $\text{L}_2$-norm is adopted in our experiments).
Then, $k$~($1\leq k \ll n$) nearest adapters will be activated and selected.
Assume that the activated adapters are $a_1', a_2',\dots,a_k'$.
According to the encoding and the activated adapters, the final prediction for $x$ is calculated through ensemble as:
\begin{equation}
    p = \frac{1}{k}\sum_{z=1}^{k} \bm{M}_z' (\bm{e}, \theta_z'),
\end{equation}
where $\bm{M}_z'$ denotes the sub-model of $a_z'$, and $\theta_z'$ denotes the parameters of $\bm{M}_z'$.

\begin{table}[t]
\centering
\setlength{\tabcolsep}{4mm}
\begin{tabular}{ll}
\toprule
\textbf{Symbol} & \textbf{Definition} \\
\midrule
$x$ & The data instance. \\
$y$ & The label of the data instance. \\
$\bm{e}$ & The encoding of the data instance. \\
$\mathcal{D}$ & The training dataset. \\
$\mathrm{En}$ & The encoder of LegoNet. \\
$a_j$ & The $j^\text{th}$ adapter of LegoNet. \\
$\bm{K}_j$ & The key of $a_j$. \\
$\bm{M}_j$ & The sub-model of $a_j$. \\
$\theta_j$ & The parameters of $\bm{M}_j$. \\
$\mathcal{D}_{a_j}$ & The set of samples that can activate $a_j$.\\  
\bottomrule
\end{tabular}
\caption{Definitions of Main Symbols.}
\label{symbol}
\end{table}

\subsection{Pre-Settings and Training}
In this section, we introduce: (1) the pre-settings for the encoder and the keys of the adapters; (2) the training process for the sub-model of each adapter.
Assume the training dataset has $N$ samples, specifically, $\mathcal{D}\coloneqq\{(x_i, y_i) \colon 1\leq i \leq N\}$,
where the subscript $i$ represents the index of the instance.

For the encoder, before being adopted in LegoNet, it should be pre-trained on external data to ensure its enough ability for representation.
The choice for the external data can be public free data, desensitized data, or non-personal data.
This setting ensures that the data will not be required to unlearn samples.
In this way, LegoNet does not need to reprocess the encoder.
In particular, data from a similar domain to the target task is a better choice for pre-training the encoder, which can effectively improve the performance of LegoNet.

For the keys of the adapters, after the encoder is pre-trained, the keys~($\bm{K}_1,\dots,\bm{K}_n$) should be initialized to have the similar distribution with the encodings of training samples in $\mathcal{D}$.
In the specific implementation of initialization, we ensure that the key distribution of adapters and the encoding distribution of training samples are the same on expectation via Monte Carlo Sampling~\cite{shapiro2003monte}.
Specifically, we uniformly sample $n$ instances (denoted as $x_1^*,\dots, x_n^*$) from $\mathcal{D}$ and adopt the encoding of $x_j^*$ to build the key of $a_j$ as:
\begin{equation}
    \label{equ:presetkey}
    \bm{K}_j=\mathrm{En}(x_j^*) + \xi_j,
\end{equation}
where $\xi_j$ represents a slight random perturbation.
$\xi_j$ is adopted to guarantee that the information of the sample is not directly recorded in LegoNet.
After initialization, the keys of adapters should be fixed.

The keys directly determine which samples the adapters will process (including training and inference).
During training, when each sample activates its $k$ nearest adapters, we record the set of samples for activating different adapters.
Specifically, we denote the set of samples that activate $a_j$ as $\mathcal{D}_{a_j}$.
Our setting for keys make $\{|\mathcal{D}_{a_j}|\}_{j=1,\dots,n}$ roughly follow the uniform distribution.
Since there are more adapters where the samples are dense in the encoding space (similarly, there are fewer adapters where the samples are sparse).
The balanced samples for each adapter are beneficial to the training quality of the sub-models, because it mitigates the situation that a sub-model is lack of enough training data.
Moreover, it also facilitates the subsequent unlearning task by avoiding that a sub-model should be re-trained by a large number of samples.
In particular, for the effect of the distribution of  $\{|\mathcal{D}_{a_j}|\}_{j=1,\dots,n}$ on unlearning efficiency, we provide a more detailed analysis in the appendix.

For the sub-models, after the keys of adapters are set, we individually train $\bm{M}_j$ by $\mathcal{D}_{a_j}$.
Specifically, for $j=1,\dots, n$, the parameters of $\bm{M}_j$ is optimized by:
\begin{equation}
    \theta_j=\argmin_\theta \sum_{(x, y)\in \mathcal{D}_{a_j}} \mathcal{L}(\bm{M}_j(\mathrm{En}(x), \theta), y),
\end{equation}
where $\mathcal{L}$ denotes the loss function.
In particular, independent training ensures that the adapter $a_j$ can only be impacted by the samples of $\mathcal{D}_{a_j}$.

\subsection{Unlearning Process}
\label{subsec:unlearn}

Unlearning requests are usually intermittent and sequential, \eg on online services.
When a training sample is requested to be unlearned, LegoNet locates and re-trains the impacted adapters to remove the sample's impact.
Specifically, for unlearning $x$, LegoNet just needs, as in the inference process, to encode $x$ as $\boldsymbol{e}=\mathrm{En}(x)$, and then we can find the $k$ nearest adapters to this sample in the encoding space.
Assume such $k$ adapters are $a_1^x, a_2^x, \dots, a_k^x$.
Because the parameters of the encoder and the keys of the adapters are fixed, the activated adapters are the impacted ones that have been trained on $x$.
During training, we have recorded the set of samples that each adapter has been trained on, \ie~$\mathcal{D}_{a_1}, \dots \mathcal{D}_{a_n}$.
For unlearning, we need to remove the unlearn sample $x$ from $\mathcal{D}_{a_z^x}$ as
\begin{equation}
    \mathcal{D}_{a_z^x}\leftarrow \{(x',y'):(x',y')\in \mathcal{D}_{a_z^x},x'\neq x\}
\end{equation}
to update the sample set for re-training the sub-model of $a_z^x$, where $z=1$ to $k$.
Specifically, the unlearning process of LegoNet is shown in \cref{alg:retrain}.
Every time LegoNet receives an unlearning request, it will perform the unlearning process once.

\begin{algorithm}[tb]
  \caption{Unlearning Process of LegoNet}
  \label{alg:retrain}
\begin{algorithmic}
    \STATE {\bfseries Components:}  Encoder $\mathrm{En}$, Adapters $\{a_1,\dots, a_n\}$
    \STATE {\bfseries Sample Records:}  $\mathcal{D}_{a_1}, \dots \mathcal{D}_{a_n}$
  \STATE {\bfseries Unlearning Data:} $x$
    \STATE Calculate the distance $\sigma(x, a_j)$, for $j=1,\dots,n$
    \STATE $\{\mathcal{D}_{a_1^x}, \dots, \mathcal{D}_{a_k^x}\}\leftarrow$ activate $k$ nearest adapters 
    \FOR{$\mathcal{D}_{a_z^x}$ {\bfseries in} $\{\mathcal{D}_{a_1^x}, \dots, \mathcal{D}_{a_k^x}\}$}
    \STATE $\mathcal{D}_{a_z^x}\leftarrow \{(x',y'):(x',y')\in \mathcal{D}_{a_z^x},x'\neq x\}$
    \STATE $\bm{M}_{a_z^x}\leftarrow $Random initial model
    \STATE Re-train $\bm{M}_{a_z^x}$ on $\mathcal{D}_{a_z^x}$
    \ENDFOR
\end{algorithmic}
\end{algorithm}

\subsection{Comparing LegoNet with SISA}
\label{subsec:discuss}
\begin{figure}[tb]
\centering
\includegraphics[width=0.9\linewidth]{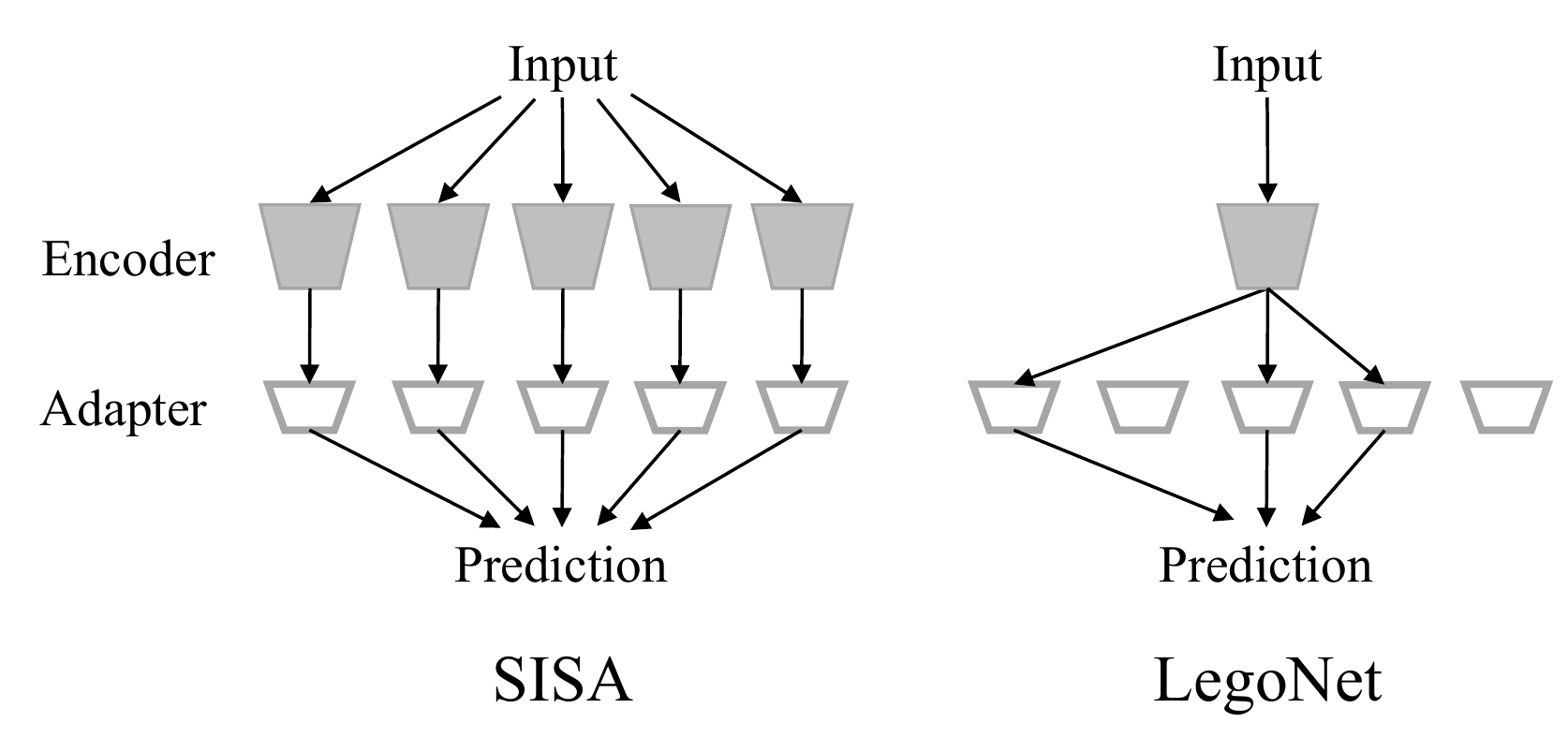}
\caption{Frameworks of SISA and LegoNet. For SISA, we set the number of shards $s=5$. For LegoNet, we set $n=5$ and $k=3$. In actual use, $k\ll n$ will be satisfied.}
\label{fig:sisaandlego}
\end{figure}

SISA, as a representative of the previous methods for exact unlearning, also aims to accelerate re-training.
In this section, we compare LegoNet with SISA to further discuss the advantages of our method.
Following the naming in this paper, in a common framework, the original neural model consists of an encoder and an adapter.
On this basic, \cref{fig:sisaandlego} shows the frameworks of SISA and LegoNet.
SISA splits the training data into $s$ disjoint shards.
Then, SISA adopts such $s$ shards to train $s$ models independently.
The architecture of each model can be viewed as an encoder connecting an adapter.
While LegoNet shares the fixed encoder for $n$ adapters.

Compared with SISA, LegoNet has higher unlearning efficiency.
On the one hand, LegoNet has few parameters that need to be re-trained.
For unlearning a sample, SISA needs to re-train an encoder and an adapter, while LegoNet needs to re-train $k$ adapters.
Since $k$ is generally small and the parameters of the encoder are significantly more than that of the adapter.
The parameters of the encoder are even much more than that of $k$ adapters.

On the other hand, while maintaining acceptable performance, LegoNet can furthermore reduce the expected number of re-training samples.
Assume the training data has $N$ samples.
For SISA, each model is expected to train $N/s$ samples.
For LegoNet, each adapter is expected to train $kN/n$ samples.
Although LegoNet needs to re-train $k$ adapters for unlearning a sample, since $k\ll n$, LegoNet can even achieve $k^2N/n<N/s$.
This is thanks to the design of the activation mechanism.
The activation mechanism of LegoNet is based on the distance in the encoding space.
Similar samples tend to activate the same adapter.
When inferring a sample, the knowledge brought from similar samples is usually more important.
Compared to SISA's randomly assigning samples to train the models, our method makes the learning of the adapter to be more focused on a small region of the encoding space. 
So our adapters can effectively maintain the performance while requiring fewer samples.

\section{Experiments}

In this section, we conduct image classification experiments to present the ability of LegoNet.
We first introduce the settings of experiments, including the datasets, baselines, and detailed settings of models.
Then we show the performance of our method for various unlearning tasks.
Moreover, we analyze the advantages of LegoNet over SISA. 
Finally, we present the characteristics of LegoNet, through extensive experiments with different hyperparameters.

\subsection{Experimental settings}
\label{sec:exp-set}
\subsubsection{Datasets}
We conduct our experiments on two standard image classification datasets: CIFAR-10 (containing 10 classes) and CIFAR-100 (containing 100 classes).
These two datasets are commonly adopted for validating the model performance for classification, and they are also frequently adopted in previous research on machine unlearning~\cite{mahadevan2021certifiable,2021AdaptiveMU,bourtoule2021machine}.
In addition, our model requires an external dataset to pre-train the encoder for representation learning. 
Here, toward the CIFAR datasets, we adopt ImageNet for pre-training the encoder~\cite{bourtoule2021machine,kornblith2019better}.
In particular, our experiments do not discuss the situation where there is no suitable data for pre-training. This question is also the direction of our future works.

\subsubsection{Unlearning Tasks}
To test the unlearning performance, our experiments follow the previous studies~\cite{golatkar2020eternal,chundawat2022zero}.
For each dataset, we conduct two kinds of unlearning tasks:
\begin{itemize}[topsep=0pt, itemsep=1pt, parsep=0pt]
    \item Random Sample Unlearning: we randomly select some training samples to be unlearned from the trained model. In the following, we name the adopted unlearning task to unlearn 100 samples as ``\textbf{Random100}''.
    \item Random Class Unlearning: we randomly select a class, and the samples in this class are requested to be unlearned. In the following, we name this task as ``\textbf{UnClass}''. 
\end{itemize}

\subsubsection{Performance Evaluation}
To identify the sample sets, we denote $D_{\text{unlearn}}$ as the set of unlearned samples, and denote $D_{\text{retain}}$ as the set of the other samples in the training data.
Moreover, we denote $D_{\text{test}}$ as the test sample set.
In the experiments, we will focus on the accuracy on $D_{\text{unlearn}}$, $D_{\text{retain}}$, $D_{\text{test}}$ to evaluate the performance of the unlearning strategies.

\subsubsection{Implementation Details of Models}
Since different frameworks may have different strategies for unlearning. 
Here, we give the setting of different methods in detail.
Specifically, in our experiments, there are three frameworks for unlearning:
\begin{itemize}[topsep=0pt, itemsep=1pt, parsep=0pt]
    \item \textbf{LegoNet}: Our model consists of an encoder and $n$ adapters. Specifically, we adopt a ResNet with 34 layers~(ResNet-34 for short, ResNet-34 is deep enough to solve CIFAR-10/100) as our encoder.
    Each adapter is set as a linear layer.
    \item \textbf{Normal}:  The neural network is consist of a ResNet-34 backbone for representation and a linear layer for classification. 
    In particular, we can think of this network as being composed of an encoder and an adapter.
    In the following, we call this ResNet Classifier as \textbf{RNC} for short.
    \item \textbf{SISA}: The model consists $s$ sub-models. Each sub-model adopts the same architecture as the above RNC.
\end{itemize}
For a fair comparison, ResNet-34s, adopted in LegoNet, RNC, and SISA, are pre-trained on ImageNet.
In addition, the unlearning strategies adopted in the experiments are:
\begin{itemize}[topsep=0pt, itemsep=1pt, parsep=0pt]
    \item \textbf{Re-Train}: Re-training the model on $D_{\text{retain}}$.
    \item \textbf{Tune}: Continue fine-tuning the trained model on $D_{\text{retain}}$~\cite{golatkar2020eternal}.
    \item \textbf{NGrad}: Continue training the trained model on $D_{\text{unlearn}}$ with opposite gradients~\cite{golatkar2020eternal}.
\end{itemize}
Specifically, ``Re-Train'' is adopted on RNC, SISA, and LegoNet.
While ``Tune'' and ``NGrad'' are approximate unlearning methods, so we only adopt them on RNC for comparison.
For more detailed settings, please refer to the appendix.

\subsection{Main Results}
\label{sec:exp-unlearn}

\begin{table*}[t]
\begin{center}
\begin{small}
\begin{adjustbox}{max width=0.98\textwidth}
\begin{tabular}{llrrrrrrrrrrrrc}
\toprule
\multicolumn{2}{c}{
\multirow{3}{*}{
\diagbox{Method}{Task}
}
} & \multicolumn{6}{c}{CIFAR-10} & \multicolumn{6}{c}{CIFAR-100} & \multirow{2}{*}{
\shortstack[c]{Overall\\Average}
}\\
\cmidrule(lr){3-8} \cmidrule(lr){9-14}
 & &
\multicolumn{3}{c}{Random100} & 
\multicolumn{3}{c}{UnClass} & 
\multicolumn{3}{c}{Random100} & 
\multicolumn{3}{c}{UnClass} \\
\cmidrule(lr){3-5} \cmidrule(lr){6-8}
\cmidrule(lr){9-11} \cmidrule(lr){12-14} \cmidrule(lr){15-15} 
& & $D_{\text{retain}}$ & $D_{\text{unlearn}}$ & $D_{\text{test}}$ & $D_{\text{retain}}$ & $D_{\text{unlearn}}$ & $D_{\text{test}}$ & $D_{\text{retain}}$ & $D_{\text{unlearn}}$ & $D_{\text{test}}$ & $D_{\text{retain}}$ & $D_{\text{unlearn}}$ & $D_{\text{test}}$ & $D_{\text{test}}$\\
\midrule
\multirow{4}{*}{\shortstack[l]{RNC}} & Origin 
& 100.0 & 100.0 & 95.97 
& 100.0 & 100.0 & 95.97
& 99.98 & 100.0 & 80.98 
& 99.98 & 100.0 & 80.98 & 88.48\\
& Re-train 
& 100.0 & 96.00 & 95.85
& 100.0  & 0.00 & 86.41
& 99.98 & 81.00 & 81.23 
& 99.99 & 0.00 & 80.33 & 85.96\\
& Tune 
& 99.98 & 97.00 & 95.88 
& 99.99 & 0.00 & 86.03
& 99.95 & 90.00 & 80.71 
& 99.94 & 0.00 & 79.71 & 85.58\\
& NGrad 
& 96.48 & 94.00 & 90.99 
& 89.54 & 0.00 & 75.88
& 92.67 & 88.00 & 73.45 
& 95.28 & 0.50 & 70.27 & 77.65\\
\midrule
\multirow{2}{*}{\shortstack[l]{SISA\\$s=10$}} &
Origin 
& 93.32 & 92.00 & 91.66 
& 92.98 & 96.35 & 91.66
& 81.78 & 77.00 & 75.84 
& 81.64 & 93.75 & 75.84 & 83.75\\
& Re-train 
& 93.14 & 91.00 & 91.56 
& 90.56 & 0.00 & 79.61
& 81.09 & 73.00 & 74.74 
& 82.25 & 0.00 & 75.42 & 80.33\\
\midrule
\multirow{2}{*}{\shortstack[l]{LegoNet\\$n=100$}} 
& Origin 
& 93.47 & 94.00 & 90.11 
& 93.19 & 96.03 & 90.11
& 87.55 & 71.00 & 72.44 
& 87.48 & 90.00 & 72.44 & 81.28\\
& Re-train 
& 93.46 & 94.00 & 89.97 
& 93.49 & 0.00 & 80.82
& 87.58 & 70.00 & 72.4 
& 87.46 & 0.00 & 71.58 & 78.69\\
\midrule
\multirow{2}{*}{\shortstack[l]{LegoNet\\$n=1000$}} 
& Origin 
& 94.66 & 94.00 & 89.45 
& 94.47 & 96.35 & 89.45
& 93.59 & 73.00 & 70.64 
& 93.49 & 98.5 & 70.64 & 80.05\\
& Re-train 
& 94.77 & 95.00 & 89.45 
& 94.59 & 0.00 & 80.31
& 93.53 & 77.00 & 70.46
& 93.49 & 0.00 & 69.75 & 77.49\\
\bottomrule
\end{tabular}
\end{adjustbox}

\end{small}
\end{center}
\caption{Accuracy (\%) on CIFAR-10/100 for Evaluating the Performance of Unlearning Strategies. The ``Origin'' rows present the results before unlearning. All LegoNets in this table set $k=10$. The ``Overall Average'' presents the average results of all unlearning tasks.}
\label{tab:main}

\end{table*}

To demonstrate the performance, we compare the accuracies of our approach on $D_{\text{retain}}$, $D_{\text{unlearn}}$, and $D_{\text{test}}$ with multiple unlearning baselines.
As shown in \cref{tab:main}, we can see that:
\begin{enumerate*}[label=(\arabic*)]
    \item Observing the performance on $D_{\text{test}}$, the accuracy of SISA and LegoNet is lower than that of RNC, which indicates that speeding up unlearning comes at a performance cost.
    \item For RNC, re-training is the reliable way to achieve exact unlearning. However, its unlearning costs too much time.
    Tune and NGrad are approximate unlearning methods. Their unlearning performance is generally judged indirectly by comparing with the accuracy of re-training on $D_{\text{retain}}$ and $D_{\text{unlearn}}$.
    The results of Tune are similar to Re-Train, but the unlearning speed is also too slow because the fine-tuning is processed on $D_{\text{retain}}$.
    In particular, NGrad unlearns very fast, but it can be found that the unlearning of NGrad is not reliable.
    In the ``UnClass'' task for CIFAR-100, the model still holds the probability of inferring the unlearned class, so the impact of this unlearned class is not completely removed.
    \item 
    Comparing SISA($s=10$) and LegoNet($n=100$), the overall test accuracy of LegoNet is slightly lower than that of SISA~(the average is 1.64\%). This is the price of the fixed encoder, but in exchange for the faster unlearning speed of LegoNet.
    As the time cost shown in \cref{fig:performance}, LegoNet($n=100$) only costs 32.65\% of the time of SISA($s=10$) for achieving unlearning.
    \item Comparing LegoNet($n=100$) and LegoNet($n=1000$), it shows how our method performs when it trades off model performance against unlearning efficiency.
    Increasing the degree of data segmentation results in higher unlearning speed but leads to a decrease in performance because the samples for training each adapter are reduced.
    \item Further comparing LegoNet($n=1000$) and SISA($s=10$), The overall test accuracy of LegoNet($n=1000$) is 2.84\% lower than that of SISA($s=10$). But LegoNet($n=100$) only costs 12.91\% of the time of SISA($s=10$) for compeleting the unlearning process.
\end{enumerate*}
Overall, LegoNet trades a small performance loss for a huge speed boost.

\pgfplotsset{
axis background/.style={fill=gallery},
grid=both,
  xtick pos=left,
  ytick pos=left,
  tick style={
    major grid style={style=white,line width=1pt},
    minor grid style=gallery,
    draw=none
    },
  minor tick num=1,
  ymajorgrids,
	major grid style={draw=white},
	y axis line style={opacity=0},
	tickwidth=0pt,
}

\begin{figure}
\centering
    \begin{tikzpicture}[scale=0.48]
	\begin{groupplot}[
	    group style={group size=2 by 1,
	        horizontal sep = 60pt,
	        }, 
	    xlabel=\huge Data Segmentation,
        ylabel=\huge Accuracy (\%),
        xtick={10,20,50,100},
        xticklabel style={align=left},
        ymajorgrids,
        major grid style={draw=white},
        y axis line style={opacity=0},
        tickwidth=0pt,
        yticklabel style={
        /pgf/number format/fixed,
        /pgf/number format/precision=5
        },
        every tick label/.append style={font=\Large},
        scaled y ticks=false,
        every axis title/.append style={at={(0.5,1)},font=\Huge},
        nodes near coords,
		every node near coord/.append style={anchor=south, font=\tiny},%
	    ]
		\nextgroupplot[
		legend style = {
		  font=\small ,
          draw=none, 
          fill=none,
          column sep = 2pt, 
          /tikz/every even column/.append style={column sep=5mm},
          legend columns = -1, 
          legend to name = grouplegend},
		title={\huge Performance}, 
		]
		\addplot[thick,color=tuatara,mark=pentagon, every node near coord/.append style={xshift=2mm}] coordinates {
          (10, 91.66)
          (20, 90.10)
          (50, 81.27)
          (100, 72.88)
        }; \addlegendentry{SISA}
         \addplot[thick,color=free_speech_aquamarine,mark=triangle*, every node near coord/.append style={anchor=north}] 
        coordinates {
          (10, 90.11)
          (20, 90.08)
          (50, 89.53)
          (100, 89.45)
        }; \addlegendentry{$\text{LegoNet}$}

        \nextgroupplot[
		title={\huge Time Cost per Unlearning}, 
		ylabel=\huge Seconds,
		]
		\addplot[thick,color=tuatara, mark=pentagon, every node near coord/.append style={xshift=2mm}]
		coordinates {
          (10, 951.15)
          (20, 493.38)
          (50, 233.52)
          (100, 146.82)
        }; 
         \addplot[thick,color=free_speech_aquamarine,mark=triangle*, every node near coord/.append style={anchor=north}] 
        coordinates {
          (10, 310.56)
          (20, 190.73)
          (50, 133.82)
          (100, 122.78)
        };

	\end{groupplot}
\node at ($(group c1r1) + (120pt, 120pt)$) {\ref{grouplegend}};
\end{tikzpicture}
    \caption{Test Accuracy (\%) and Unlearning Time of SISA and LegoNet under Different Degrees of Data Segmentation. Here, the dataset is CIFAR-10. The unlearning task is ``Random100''. The ``Time Cost per Unlearning'' shows the average time of unlearning 100 times.}
    \label{fig:performance}
\end{figure}
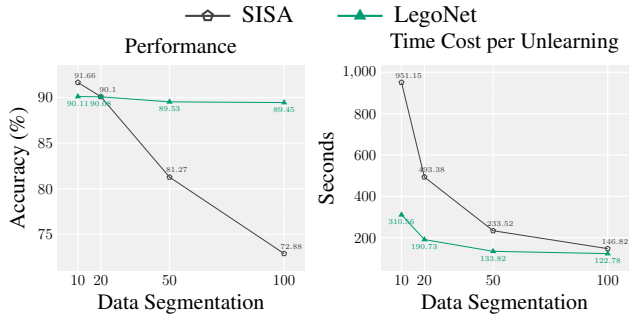

To further demonstrate the advantages of LegoNet in unlearning acceleration, we compare SISA and LegoNet under different degrees of data segmentation on the ``Random100'' task of CIFAR-10.
For SISA, the degree of data segmentation is the value of $s$.
For LegoNet, we set $k=10$ and the degree of data segmentation is the value of $n/k$.
In particular, when $s=n/k$, the comparison of SISA and LegoNet is relatively fair because the amount of data for re-training each sub-model is expected to be equal.
As shown in \cref{fig:performance}, we can see that:
\begin{enumerate*}[label=(\arabic*)]
\item As the degree of data segmentation increases, the performance of SISA drops dramatically. This indicates that the SISA has a very limited efficiency for unlearning.
\item The unlearning speed of LegoNet is faster than that of SISA when the degree of data segmentation is the same.
\item LegoNet's performance is well maintained, when $n$ is increased from 100 to 1000.
Moreover, the accuracy of LegoNet($n=1000$) far outperforms that of SISA($s=100$), which indicates that LegoNet can support a higher degree of data segmentation than SISA. 
Therefore, LegoNet is more suitable when there are higher rigid requirements for unlearning speed.
\end{enumerate*}

The reason why our method can achieve faster unlearning is very intuitive.
On the one hand, the number of parameters that LegoNet needs to be re-trained is much smaller than that of SISA.
This is because the parameters of the encoder are thousands of times more than that of an adapter.
Specifically, for CIFAR-10, LegoNet($k=10$) needs to re-train 51.2 thousand parameters.
While $\text{SISA}$ needs to re-train 3.68 million parameters.
On the other hand, due to the simple architecture of the adapter, the number of epochs required for training the adapter is far less than that for training an encoder.
Overall, while maintaining acceptable accuracy, LegoNet achieves faster exact unlearning.
Therefore, considering the comprehensive ability in terms of speed, accuracy, and unlearning, LegoNet outperforms other baselines.

\subsection{Ablative Analysis}
\label{subsec:compare}

\begin{table*}[t]
\setlength{\tabcolsep}{6mm}
\centering
\begin{adjustbox}{max width=0.98\textwidth}
\begin{tabular}{lcccccccc}
\toprule

Dataset &
\multicolumn{4}{c}{CIFAR-10} & 
\multicolumn{4}{c}{CIFAR-100} \\
\cmidrule(lr){2-5} \cmidrule(lr){6-9}
Segmentation & 10 & 20 & 50 & 100 & 10 & 20 & 50 & 100\\
\midrule
SISA & \textbf{91.66} & \textbf{90.10} & 81.27 & 72.88 & \textbf{75.84} & \textbf{72.94} & 64.20 & 47.15\\
FixSISA & 88.32 & 87.00 & 85.23 & 83.76 & 69.39 & 67.42 & 64.36 & 62.44\\
$\text{LegoNet}_{k=1}$ & 87.90 & 87.46 & 86.40	& 84.85 & 67.14 & 64.85 & 63.44 & 61.82\\
$\text{LegoNet}_{k=10}$ & 90.11 & 90.08 & \textbf{89.53} & \textbf{89.45} & 72.44 & 72.08 & \textbf{71.10} & \textbf{70.64}\\
\bottomrule
\end{tabular}
\end{adjustbox}
\caption{Accuracy (\%) in Ablation Experiments. For SISA and FixSISA, the number in ``Segmentation'' denotes the value of $s$, For LegoNet, the number in ``Segmentation'' denotes the value of $n/k$. }
\label{tab:abla}
\end{table*}

To study how LegoNet outperforms SISA, \ie what kinds of differences contribute to the improvements, we ablate them separately.
When LegoNet sets $k=1$, denoted as $\text{LegoNet}_{k=1}$, each training sample will only be assigned to one sub-model.
Set all sub-models of SISA to share an encoder, and this encoder is fixed during training.
We call this model FixSISA.
Then, the only difference between FixSISA and $\text{LegoNet}_{k=1}$ is whether to adopt distance-based data segmentation and sub-model integration.
For the above models, we conduct experiments under the condition that the expectation of the number of samples allocated to a single sub-model is equal~(\ie $s=n/k$).

As shown in \cref{tab:abla}, we can see that:
\begin{enumerate*}[label=(\arabic*)]
    \item The performance of SISA drops seriously when $s$ increases to 50 and 100. This is because the data accepted by each sub-model is too sparse, which greatly increases the learning difficulty. 
    Moreover, compared with SISA, FixSISA performs better when $s=50, 100$.
    This indicates that a large number of sub-model parameters makes overfitting more likely to occur, thus affecting the results of the ensemble.
    \item Compared with FixSISA, $\text{LegoNet}_{k=1}$ changes the strategy of assigning data and integrating. From the results, the ability of $\text{LegoNet}_{k=1}$ is approximately the same as that of FixSISA.
    But $\text{LegoNet}_{k=1}$ only adopts one sub-model for inference while FixSISA adopts its all sub-models.
    It verifies the rationality of our adapter activation mechanism, assigning data based on similarity.
    The sub-model of LegoNet to infer a sample is trained on the set of similar samples, which means LegoNet only adopts the knowledge of similar samples. While SISA actually uses the knowledge of all the samples. From the comparison between $\text{LegoNet}_{k=1}$ and FixSISA, we can see that similar samples are more important for inference.
    \item LegoNet has higher freedom in assigning samples because one sample can be assigned to $k$ sub-models. Comparing $\text{LegoNet}_{k=10}$ with $\text{LegoNet}_{k=1}$, it shows that the overall capability of LegoNet is greatly improved when $k$ increases.
\end{enumerate*}

\subsection{LegoNet with Different Hyperparameters}
\label{subsec:characteristics}

\pgfplotsset{
axis background/.style={fill=gallery},
grid=both,
  xtick pos=left,
  ytick pos=left,
  tick style={
    major grid style={style=white,line width=1pt},
    minor grid style=gallery,
    draw=none
    },
  minor tick num=1,
  ymajorgrids,
	major grid style={draw=white},
	y axis line style={opacity=0},
	tickwidth=0pt,
}

\begin{figure}
\centering
    \begin{tikzpicture}[scale=0.33]
	\begin{groupplot}[
	    group style={group size=3 by 1,
	        horizontal sep = 48pt,
	        }, 
	    xlabel=\huge Hyperparameter,
        ylabel=\huge Accuracy (\%),
        xtick={1,2,3,4},
        xticklabel style={align=left},
        ymajorgrids,
        major grid style={draw=white},
        y axis line style={opacity=0},
        tickwidth=0pt,
        yticklabel style={
        /pgf/number format/fixed,
        /pgf/number format/precision=5
        },
        every tick label/.append style={font=\Large},
        scaled y ticks=false,
        every axis title/.append style={at={(0.5,1)},font=\Huge},
        nodes near coords,
		every node near coord/.append style={anchor=south, font=\scriptsize},%
	    ]
		\nextgroupplot[
		xticklabels={{$n=100$\\ $k=10$}, {$n=200$\\ $k=10$}, {$n=500$\\ $k=10$}, {$n=1000$\\ $k=10$}},
		legend style = {
		  font=\scriptsize ,
          draw=none, 
          fill=none,
          column sep = 2pt, 
          /tikz/every even column/.append style={column sep=5mm},
          legend columns = -1, 
          legend to name = grouplegend},
		title={Fix $k=10$}, 
		]
		\addplot[thick,color=tuatara,mark=pentagon] coordinates {
          (1, 90.11)
          (2, 90.08)
          (3, 89.53)
          (4, 89.45)
        }; \addlegendentry{CIFAR-10}
         \addplot[thick,color=free_speech_aquamarine,mark=triangle*] coordinates {
          (1, 72.44)
          (2, 72.08)
          (3, 71.1)
          (4, 70.64)
        }; \addlegendentry{CIFAR-100}

        \nextgroupplot[
        xticklabels={{$n=100$\\ $k=1$}, {$n=100$\\ $k=2$}, {$n=100$\\ $k=5$}, {$n=100$\\ $k=10$}},
        title={Fix $n=100$}]
		\addplot[thick,color=tuatara,mark=pentagon] coordinates {
          (1, 84.85)
          (2, 87.66)
          (3, 89.72)
          (4, 90.11)
        };
         \addplot[thick,color=free_speech_aquamarine,mark=triangle*] coordinates {
          (1, 61.82)
          (2, 66.96)
          (3, 71.64)
          (4, 72.44)
        };
        
       \nextgroupplot[
       xticklabels={{$n=10$\\ $k=1$}, {$n=20$\\ $k=2$}, {$n=50$\\ $k=5$}, {$n=100$\\ $k=10$}},
       title={Fix $n/k = 10$},
       every axis title/.append style={at={(0.5,0.96)},font=\Huge}]
		\addplot[thick,color=tuatara,mark=pentagon] coordinates {
          (1, 87.9)
          (2, 88.84)
          (3, 89.94)
          (4, 90.11)
        };
         \addplot[thick,color=free_speech_aquamarine,mark=triangle*] coordinates {
          (1, 67.14)
          (2, 69.69)
          (3, 72.31)
          (4, 72.44)
        };

	\end{groupplot}
\node at ($(group c1r1) + (250pt, 120pt)$) {\ref{grouplegend}};
\end{tikzpicture}
    \caption{Test Accuracy (\%) of LegoNet with Different Hyperparametrics.}
    \label{fig:hyperparam}
\end{figure}
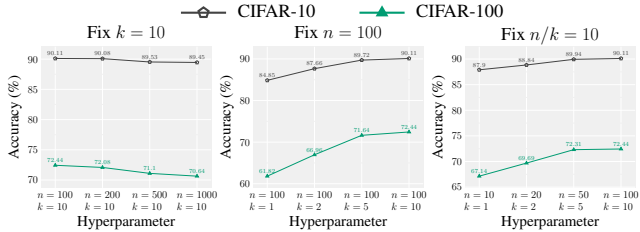

To show the characteristics of LegoNet, we evaluate LegoNet's ability with different hyperparameters.
First, we discuss the results of LegoNet by fixing $k$, $n$, and $n/k$, respectively.
As shown in \cref{fig:hyperparam}, we can see that:
\begin{enumerate*}[label=(\arabic*)]
    \item When fixing $k=10$, 
    the increase of $n$ will lead to a decrease in model performance, because the data received by each sub-model is significantly reduced. 
    However, LegoNet maintains performance well. Comparing $n=100$ and $n=1000$ the accuracy only decreases by 0.66\% in CIFAR-10 and 1.80\% in CIFAR-100.
    \item When fixing $n=100$, 
    the increase of $k$ will lead to increasing the model performance.
    There are two reasons for this.
    On the one hand, the samples for training each sub-model are increased.
    On the other hand, more sub-models have joined the ensemble of results.
    \item When fixing $n/k=10$, LegoNet's performance improves as $n$ and $k$ are increased synchronously.
    The expected number of samples received by each sub-model is unchanged. So the performance improvement here is brought about by integrating more sub-models.
\end{enumerate*}
Overall, the performance of LegoNet is stable when $k$ is large enough. In practical tasks, a large increase on $n$ can greatly improve the unlearning speed of LegoNet with a small performance loss. 
When performance needs to be increased, we can try to increase $k$.
Based on the adjustment of $n$ and $k$, LegoNet can intuitively trade off the unlearning speed against the model performance.

\pgfplotsset{
axis background/.style={fill=gallery},
grid=both,
  xtick pos=left,
  ytick pos=left,
  tick style={
    major grid style={style=white,line width=1pt},
    minor grid style=gallery,
    draw=none
    },
  minor tick num=1,
  ymajorgrids,
	major grid style={draw=white},
	y axis line style={opacity=0},
	tickwidth=0pt,
}

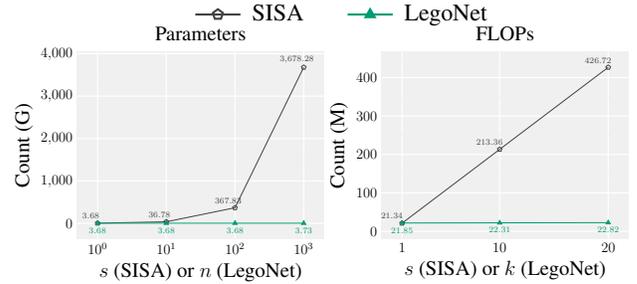
\begin{figure}
\centering
    \begin{tikzpicture}[scale=0.45]
	\begin{groupplot}[
	    group style={group size=2 by 1,
	        horizontal sep = 48pt,
	        }, 
	    width=0.5\textwidth,
	    height=0.4\textwidth,
	    xlabel=\huge Data Segmentation,
        ylabel=\huge Counting,
        xticklabel style={align=left},
        ymajorgrids,
        major grid style={draw=white},
        y axis line style={opacity=0},
        tickwidth=0pt,
        yticklabel style={
        /pgf/number format/fixed,
        /pgf/number format/precision=5
        },
        every tick label/.append style={font=\Large},
        scaled y ticks=false,
        every axis title/.append style={at={(0.5,1)},font=\Huge},
        nodes near coords,
		every node near coord/.append style={anchor=south, font=\scriptsize},%
	    ]
		\nextgroupplot[
        ylabel=\huge Count~(G),
        xlabel=\huge $s$~(SISA) or $n$~(LegoNet),,
		legend style = {
		  font=\small ,
          draw=none, 
          fill=none,
          column sep = 2pt, 
          /tikz/every even column/.append style={column sep=5mm},
          legend columns = -1, 
          legend to name = grouplegend},
		title={\huge Parameters}, 
		xmode=log,
		]
		\addplot[thick,color=tuatara,mark=pentagon,every node near coord/.append style={xshift=-2mm}] coordinates {
          (1, 3.678278656)
          (10, 36.782786560)
          (100, 367.8278656)
          (1000, 3678.278656)
        }; \addlegendentry{SISA}
         \addplot[thick,color=free_speech_aquamarine,mark=triangle*, every node near coord/.append style={anchor=north}] 
        coordinates {
          (1, 3.678278656)
          (10, 3.678739456)
          (100, 3.683347456)
          (1000, 3.729427456)
        }; \addlegendentry{$\text{LegoNet}$}

        \nextgroupplot[
        ylabel=\huge Count~(M),
		title={\huge FLOPs}, 
		xtick={1,10,20},
		xlabel=\huge $s$~(SISA) or $k$~(LegoNet),
		]
		\addplot[thick,color=tuatara,mark=pentagon, every node near coord/.append style={xshift=-3mm}]
		coordinates {
          (1,21.335872)
          (10, 213.358720)
          (20, 426.717440)
        }; 
         \addplot[thick,color=free_speech_aquamarine,mark=triangle*, every node near coord/.append style={anchor=north}] 
        coordinates {
          (1, 21.847872)
          (10, 22.308672)
          (20, 22.820672)
        };

	\end{groupplot}
\node at ($(group c1r1) + (120pt, 110pt)$) {\ref{grouplegend}};
\end{tikzpicture}
    \caption{Count of parameters and FLOPs for SISA and LegoNet. In ``FLOPs'', LegoNet sets $n=1000$. The FLOPs of LegoNet contain the computation of network inference and adapter activation.}
    \label{fig:paramsflops}
\end{figure}

Moreover, we also present the count of parameters and FLOPs of LegoNet with different hyperparameters. To intuitively show our advantages, we still choose to compare with SISA.
As shown in \cref{fig:paramsflops}, we can see that: \begin{enumerate*}[label=(\arabic*)]
\item When increasing $n$, the parameters of LegoNet change little. Because the number of parameters that an adapter has is usually small. So LegoNet requires less memory space.
\item When increasing $k$, the FLOPs of LegoNet change little. Because the encoder that needs larger FLOPs only requires to infer once.
Furthermore, LegoNet generally sets a small $k$.
\item The FLOPs of our method are significantly lower than those of SISA, which indicates the inference of LegoNet is faster.
\end{enumerate*}

The advantages in the number of parameters and the amount of computation allow our method to be more easily deployed in practical applications.
For example, the providers of online services can switch to our model at a low cost, without the need to upgrade equipment.

\section{Conclusion}
In this paper, we present a novel network architecture, namely \textit{LegoNet} to fast and completely erase the impact of specific training samples from a deep neural network.
Compared with the original model, our method only adds a very small number of additional parameters and inference calculations.
Moreover, our method has great potential. When pre-training methods are improved, our method will also benefit and achieve better performance.
From a research perspective, our method provides a new idea to speed up unlearning by both reducing the number of parameters and samples involved in the re-training process.
According to this idea, the specific implementation scheme of our proposed method still has a large space for improvement.
For example, the improvement of the encoder, a better activation mechanism for the adapters, etc.
Furthermore, because LegoNet has a strong ability to control the impact of samples, in addition to re-training, there may be more clever strategies to achieve unlearning.
We believe these future studies will be interesting and have important implications for society.

\bibstyle{aaai23}
\bibliography{unlearning}

\end{document}